%% file: main.tex
\documentclass[12pt]{article}
\usepackage[margin=1in]{geometry}
\usepackage{amsmath,amssymb,amsfonts}
\usepackage{graphicx}
\usepackage{setspace}
\usepackage{authblk}
\usepackage{cite}
\usepackage{amsmath,amssymb,amsfonts}
\usepackage{algorithmic}
\usepackage{textcomp}
\usepackage{xcolor}
\usepackage{enumitem}
\usepackage{multirow,makecell}
\usepackage{tabularx}
\usepackage{booktabs}
\usepackage{multirow}
\usepackage{graphicx}
\usepackage{tikz}
\usetikzlibrary{shapes.geometric, arrows}
\usepackage{varwidth}
\usepackage{longtable}

\def\BibTeX{{\rm B\kern-.05em{\sc i\kern-.025em b}\kern-.08em
    T\kern-.1667em\lower.7ex\hbox{E}\kern-.125emX}}

\begin{document}

\title{\textbf{VerLM: Explaining Face Verification Using Natural Language}}

\author[1,*]{Syed Abdul Hannan}
\author[1,*]{Hazim Bukhari}
\author[1]{Thomas Cantalapiedra}
\author[1]{Eman Ansar}
\author[1]{Massa Baali}
\author[1]{Rita Singh}
\author[1]{Bhiksha Raj}

\affil[1]{Carnegie Mellon University, Pittsburgh, USA}
\affil[*]{Equal Contribution}

\date{}

\maketitle

\begin{abstract}
\normalsize
\input{documents/abstract}
\end{abstract}

\section{Introduction}

\input{documents/intro}

\section{Explainable Face Verification}
\input{documents/lit_rev}

\section{Explainable Face Verification Dataset}
\input{documents/dataset}

\section{Model}
\input{documents/model}

\section{Training}
\input{documents/training}

\section{Results}
\input{documents/results}

\section{Ablations}
\input{documents/ablations}

% \section{Hallucination and Mode Collapse}
% \input{documents/hallucination}

\section{Conclusion}
\input{documents/conclusion}

\newpage
\bibliographystyle{IEEEtran}
\bibliography{documents/refs}

\newpage
\section{Appendix}
\input{documents/appendix}

\end{document}

%% file: documents/abstract.tex
Face verification systems have seen substantial advancements; however, they often lack transparency in their decision-making processes. In this paper, we introduce an innovative Vision-Language Model (VLM) for Face Verification, which not only accurately determines if two face images depict the same individual but also explicitly explains the rationale behind its decisions. Our model is uniquely trained using two complementary explanation styles: (1) concise explanations that summarize the key factors influencing its decision, and (2) comprehensive explanations detailing the specific differences observed between the images. We adapt and enhance a state-of-the-art modeling approach originally designed for audio-based differentiation to suit visual inputs effectively. This cross-modal transfer significantly improves our model's accuracy and interpretability. The proposed VLM integrates sophisticated feature extraction techniques with advanced reasoning capabilities, enabling clear articulation of its verification process. Our approach demonstrates superior performance, surpassing baseline methods and existing models. These findings highlight the immense potential of vision language models in face verification set up, contributing to more transparent, reliable, and explainable face verification systems.

%% file: documents/intro.tex
Face verification systems have experienced significant advancements in recent years, with deep learning models achieving high accuracy in identifying whether two facial images belong to the same individual \cite{Schroff_2015_CVPR, Taigman_2014_CVPR} allowing them to be used in Video surveillance and criminal identification. Gaining insight into why a decision was made by these systems allows for detecting biases and mitigating them easily and leads to more trust in decisions made by these systems, current approaches for explaining decisions depend on saliency heatmaps \cite{williford2020explainable}.

%Talk about VLMs and reasoning
Large Language Models (\textbf{LLM}) have been shown to have an emergent ability in the form of reasoning over the text modality \cite{zhang2022automatic, chowdhery11palm}, later work in the Vision Language Model (\textbf{VLM}) and Audio Language Model (\textbf{ALM}) domains such as \cite{huang2023language, liu2023visual} and \cite{rubenstein2023audiopalm, tangsalmonn} has extended reasoning abilities to different modalities. 

Using VLM's language ability to output reasoning for it's decision is valuable for understanding why the decision was made, unlike previous examples where a single input is provided our approach requires dual input to allow the model to compare and reason over them at the same time, for this we take inspiration from \cite{deshmukh2025adiff} where the authors explored using reasoning in explaining differences using dual audio input.

We introduce a novel Vision-Language Model (VLM) specifically designed for face verification. Our model not only determines whether two face images depict the same individual, but also provides explicit explanations for its decisions. This is achieved through a unique training approach that utilizes two complementary explanation styles: concise explanations that highlight the key factors influencing the decision and comprehensive explanations that detail the specific differences observed between the images.

Our approach leverages a state-of-the-art modeling framework originally developed for audio-based differentiation \cite{deshmukh2025adiff}, which we adapt and enhance to effectively process visual input. By integrating image feature extraction techniques with Large Language model (\textbf{LLM}) reasoning capabilities, our VLM enables a clear articulation of its verification process, offering a substantial step forward in the development of transparent and reliable face verification systems.

In this paper, our main contributions are as follows: 
\begin{itemize}
  \item Introduce the explainable face verification task which aims to provide a natural language
explanation for the differences between two facial images. To train and evaluate models, we created Explainable Face Verification Dataset, where the difference explanation is generated by LLM using VLM annotated captions and later verified by human annotators.
To mimic human explanations, the dataset contains two tiers of explanation: (1) concise explanations that summarize the key factors influencing its decision and (2) comprehensive explanations detailing the specific differences observed between the images.
  \item We introduce the Verification Language Model (\textbf{VerLM}) model.
  \item Under the proposed framework, we conduct several ablation studies to understand the impact of cross-projection, language model scaling, and image-grounding. Our
findings reveal that our approach outperform the baseline, in addition our approach benefits from scaled model size.
\end{itemize}

%% file: documents/lit_rev.tex
The emergence of Vision-Language Models (VLMs) has fundamentally transformed multimodal understanding, with Liu et al.'s groundbreaking work on visual instruction tuning demonstrating how large language models can be effectively adapted to understand and reason about visual content \cite{liu2023visual}. Their LLaVA framework established the feasibility of integrating sophisticated visual understanding with natural language generation through carefully designed instruction-following paradigms, paving the way for more specialized applications in computer vision tasks that require both perception and linguistic reasoning.

\textbf{Cross-modal explanation approaches} have gained significant traction, particularly in the audio domain where pioneering work has established effective methodologies for explaining perceptual differences using natural language. Deshmukh et al. introduced ADIFF, a novel Audio Language Model that explains differences between audio samples through natural language descriptions \cite{deshmukh2025adiff}. Their architecture employed a unique cross-projection mechanism combined with separator tokens to highlight distinguishing features between audio pairs, demonstrating that structured difference computation substantially outperforms naive feature concatenation approaches. Complementing this work, Elizalde et al. developed CLAP (Contrastive Language-Audio Pre-training), which established robust methods for learning audio concepts from natural language supervision through contrastive learning in shared embedding spaces\cite{elizalde2023clap}. These audio-centric approaches provided crucial architectural insights and demonstrated the viability of conditioning language models on non-textual inputs to generate meaningful explanations.

\textbf{Image difference analysis} has been advanced significantly by Hu et al.'s OneDiff model, which introduced specialized architectures for general image difference captioning \cite{hu2024onediffgeneralistmodelimage}. Their work emphasized the critical importance of explicit difference modeling through dedicated ``Visual Delta Modules,'' showing substantial performance improvements over baseline approaches that simply concatenate image features. OneDiff demonstrated that successful image difference explanation requires architectural components specifically designed to capture and articulate distinguishing characteristics, rather than relying on implicit difference detection within standard vision-language frameworks. However, their approach focused primarily on general image differences rather than domain-specific applications, leaving substantial room for specialized optimizations in areas requiring fine-grained analysis such as face verification.

\textbf{Explainable face recognition} has traditionally relied on post-hoc explanation methods, primarily through saliency-based visualizations and heatmap generation. DeAndres-Tame et al. explored early integration of explainability in face recognition through interactive natural language processing, highlighting the limitations of existing approaches that provide visual explanations without rich contextual understanding \cite{deandres2024pixels}. Current explainable face recognition systems typically employ visualization-based methods that highlight important facial regions but fail to provide the detailed, human-interpretable explanations that natural language can offer, particularly for understanding specific facial similarities and differences that drive verification decisions.

\textbf{Research gaps and motivation} emerge clearly from this landscape. While general VLMs have demonstrated impressive capabilities, their application to specialized tasks like face verification requires domain-specific architectural adaptations that have not been systematically explored. Existing image difference models like OneDiff lack the fine-grained analysis capabilities required for face verification, where subtle facial features can be crucial for accurate identification. Furthermore, the successful cross-projection mechanisms demonstrated in audio domains have not been adapted and evaluated for visual face verification tasks, representing a significant opportunity for cross-modal knowledge transfer. Current approaches also typically provide single-level explanations, whereas human understanding often benefits from both high-level summaries and detailed analyses—a multi-tier approach that remains largely unexplored in face verification contexts.

Our proposed VerLM addresses these identified gaps by adapting successful cross-projection mechanisms from audio-based difference explanation to the visual domain, while incorporating domain-specific face encoders and implementing a novel dual-tier explanation system. This approach uniquely combines high-accuracy face verification with human-interpretable explanations, contributing to more transparent and trustworthy face recognition systems that can articulate their reasoning processes in natural language.

%% file: documents/dataset.tex
We chose to use a subset of the VGGFace2 dataset \cite{Cao18}, which contains 3.31 million images of 9,131 individuals, with a wide variety of poses and orientations.  We created 2 datasets, each with 79,771 pairs of images and 7,689 instances of matching faces.  We used Llama 2 to generate the descriptions for each pair of images.  Dataset 1 had an average training description length of 53.35 words, and dataset 2 had more comprehensive training descriptions with an average length of 121.64 words.  Statistics on our datasets can be found in Table \ref{tab:dataset_stats}.

We began by using the Llama 3.2 VLM to generate descriptions for each face in the dataset.  We then took random pairs of faces and assigned them labels (match or no match).  Finally, we used Llama 3.2 again to compress the descriptions for the faces in each pair into one paragraph describing similarities and differences between the individuals in the images. Additionally, with access to the label, it was prompted to write an explanation about whether the images depicted the same person.  Below are examples of descriptions generated for each dataset.

Descriptions from Dataset 1:
\begin{itemize}
    \item Faces Match: \\ \begin{center}\includegraphics[scale=0.25]{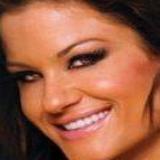} \includegraphics[scale=0.25]{images/dataset_images/0094_01.jpg}\end{center} the two facial images exhibit a high degree of similarity in key identifying features. both images depict a woman with a warm, tan complexion, almond-shaped dark brown eyes, and full eyebrows. the minor difference lies in the level of detail regarding facial hair.
    \item Faces Don't Match: \\ \begin{center}\includegraphics[scale=0.25]{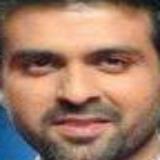} \includegraphics[scale=0.25]{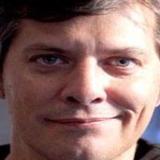}\end{center} the two facial images share some matching features, including almond-shaped eyes, straight nose, high cheekbones, strong jawline, and warm skin tone. however, notable differences in eye color, hair color and style, and facial hair suggest they do not depict the same individual.
\end{itemize}

Descriptions from Dataset 2:
\begin{itemize}
    \item Faces Match: \\ \begin{center}\includegraphics[scale=0.25]{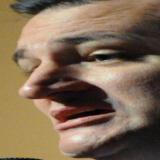} \includegraphics[scale=0.25]{images/dataset_images/0523_01.jpg}\end{center}upon analyzing the two facial images, i have identified several matching features that suggest they belong to the same individual. both images share a similar skin tone, facial structure, jawline, and nose shape. the dark hair color and style, as well as the eyebrow shape and thickness, are also consistent between the two images. however, i have noticed a slight difference in the hair texture and a minor variation in the eye color, which could be attributed to lighting conditions or minor grooming changes. despite these minor differences, the overall facial features and structure are remarkably similar, leading me to conclude that these images represent the same individual.
    \item Faces Don't Match: \\ \begin{center}\includegraphics[scale=0.25]{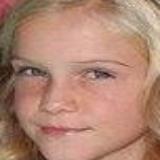} \includegraphics[scale=0.25]{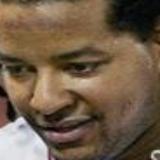}\end{center} based on the provided descriptions, the two facial images do not match. key differences include the skin tone (fair and smooth in person 1 vs. light to medium brown in person 2), facial structure (slightly pointed chin in person 1 vs. oval with a slightly rounded chin in person 2), eye color and shape (pale blue almond-like eyes in person 1 vs. dark brown almond-shaped eyes in person 2), hair color and style (light blonde wavy hair in person 1 vs. dark brown dreadlocks in person 2), facial hair (none in person 1 vs. mustache and goatee in person 2), and eyebrow shape and thickness (thin and light brown in person 1 vs. thick and dark in person 2). these significant differences indicate that the images do not belong to the same individual.
\end{itemize}

\begin{table}[ht]
    \centering
    \begin{tabular}{c|c|c|c|c} 
             & Average & Median & Max  & Vocab \\ \hline
            Dataset 1 - Train & 53.35 & 54 & 127 & 4341 \\ 
            Dataset 1 - Test & 53.71 & 54 & 124 & 1727 \\ \hline
            Dataset 2 - Train & 121.64 & 121 & 325 & 6008 \\ 
            Dataset 2 - Test & 122.19 & 121 & 263 & 2664 \\ \hline
           \end{tabular}
    \vspace{4pt}
    \caption{Description Length Statistics}
    \label{tab:dataset_stats}
\end{table}

\tikzstyle{startstop} = [rectangle, rounded corners, minimum width=3cm, minimum height=1cm,text centered, draw=blue]
\tikzstyle{io} = [trapezium, trapezium left angle=70, trapezium right angle=110, minimum width=3cm, minimum height=1cm, text centered, draw=black, fill=blue!30]
\tikzstyle{process} = [rectangle, minimum width=3cm, minimum height=1cm, text centered, draw=black, fill=orange!30]
\tikzstyle{decision} = [diamond, minimum width=3cm, minimum height=1cm, text centered, draw=black, fill=green!30]
\tikzstyle{arrow} = [thick,->,>=stealth]
\begin{center}
\begin{tikzpicture}[node distance=3cm]
\node (start) [startstop] {\begin{varwidth}{10em}Use Llama 3.2 VLM to give descriptions to all images\end{varwidth}};
\node (pairs) [startstop, below of=start] {\begin{varwidth}{10em}Take random pairs of images and assign labels\end{varwidth}};
\node (llama) [startstop, below of=pairs] {\begin{varwidth}{10em}Create a paragraph comparing the images using Llama 3.2\end{varwidth}};
\draw [arrow] (start) -- (pairs);
\draw [arrow] (pairs) -- (llama);
\end{tikzpicture}
\end{center}

%% file: documents/model.tex
In this section, we describe our proposed vision-language model, VerLM, designed for an end-to-end, explainable face verification task using natural language. The VerLM model accepts three distinct inputs: two images (image 1 and image 2) and a user-provided textual prompt. The output generated by the model is free-form text based on these inputs. 

The VerLM architecture comprises six primary components: (1) an Image Encoder, (2) an Image Projection Layer, (3) a Text Embedder, (4) a Text Projection Layer, (5) a Cross-Projection Layer, and (6) a Decoder-only Language Model.

\textbf{Image Encoder: } The Image Encoder primarily used in our experiments for extracting high-level facial representations is the VGG Face model, which achieves an accuracy of 99.65\% on the Labeled Faces in the Wild (LFW) dataset. We also provide an ablation study comparing performance when substituting the VGG Face encoder with another prominent face verification encoder, CASIA-WebFace, which achieved an accuracy of 99.05\% on the LFW dataset. Detailed results and comparative analysis from these experiments are presented in the Ablations section.

\textbf{Image Projection Layer: } This projeccton layer is used to project the face embeddings from the image encoder to the latent space of the cross projection layer and the language model. The image projection layer converts the single embedding $[b,h]$ into a sequence of latent tokens $[b,s,d]$. where $b$ is batch size, $h$ is the hidden dimension layer from the final output layer of the image encoder, $s$ is the sequence token length, and $d$ is the dimension for the input of the decoder only language model. 
To do this, we first project the hidden dimension to a larger hidden dimension $k$ where $k = s*d$. This is then followed by concatenating a learnable constant, resulting in $[s+c, d]$. Finally, this output is then passed through a transformer which is then followed by clipping of the learnable constant output $c$. The final output shape from this layer is $[b, s, d]$. This type of projection architecture is shown to perform well for prefix-tuning architectures.

\textbf{Text Embedder: } The text embedder here that is used is found from the tokenizer used in the decoder language model. In this case, we mainly employ gpt 2 based tokenizer as our text embedder. The output shape from the text embedder is $[b,t,d]$ where $b$ is the batch size, $t$ is the prompt len of the model, and $d$ is the dimension for the input of the decoder only language model.

\textbf{Text Projection Layer: } The text projection layer is very similar to the image projection layer without the converting of single embedding into a sequence of latent tokens. It is basically just concatenating the embedding from the text embedding $[b,t,d]$ with a learnable constant $c$ to make the inermediate layer $[b,t+c,d]$ and then put through a transformer layer which is then followed by the clipping of the learnable constant $c$ to get the final dimension as $[b,t,d]$. 

\textbf{Cross Projection: } Following the esteemed paper \cite{deshmukh2025adiff}, we employed a cross-projection layer to improve the models ability to highlight differences \cite{deshmukh2025adiff}. This is done by first concatenating the two image projection layer outputs along with a separator token which is the EOS token of the model. The resulting shape of this concatenation is $[b,2*s + 1,d]$. This is then further concatenated with the text projection layer to give the new dimension as $[b, 2*s + t + 1, d]$. This is then put through a similar transformer layer as the text projection layer with the clipping of the learnable constant.

\textbf{Language Model: } This is a decoder only language model which is used to autoregressively generate text conditioned on the output of the cross projection layer. We conduct various ablations with the family of gpt2 based decoder language models due to compute limitations. The model takes in the output of the cross projection layer to get the description or answer autoregressively.

\begin{figure}[htbp]
    \centering
    \includegraphics[width=1\textwidth]{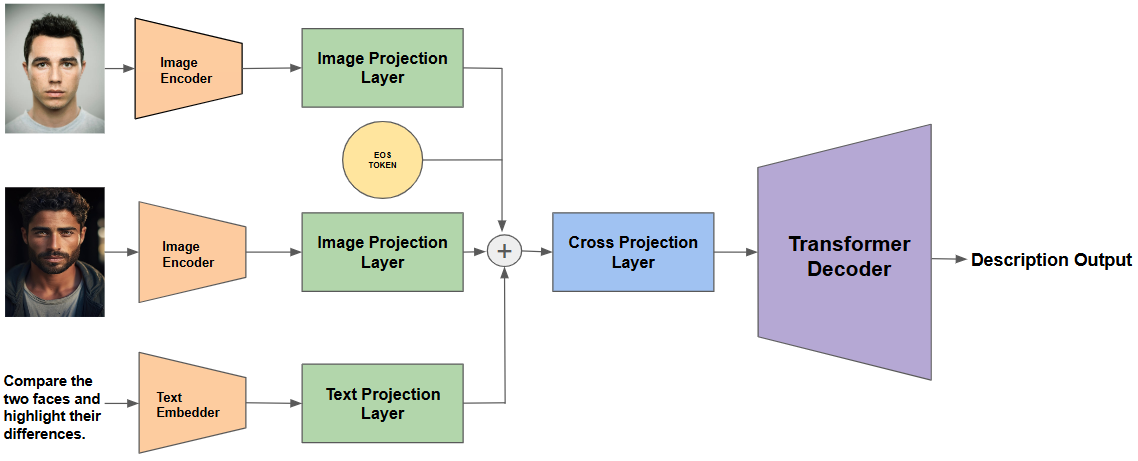}
    \caption{VerLM takes in two images and text prompt as inputs to the model. It processes the images through an image encoder and then through the image projection layer. The text is put through a text embedder followed by a text projection layer. The two encoded images are concatenated with an separator token in between and then further concatenated with the text projection layer output. This is then put through a cross-projection layer. Finally, the output of the cross-projection layer is input to the decoder model which autoregressively generates the description}
    \label{fig:VerLM_diagram}
\end{figure}

%% file: documents/training.tex
\subsection*{5.1 Training Stages}

We train the model in three distinct stages to ensure that visual features are effectively integrated into the language model. These stages are: (1) Unimodal Training, (2) Multimodal Mapper Training, and (3) End-to-End Fine-tuning.

\textbf{1. Unimodal Training:} In this stage, we independently pretrain the verification encoder and the language model on tasks specific to their respective modalities. For the vision encoder, we experiment with two architectures: VGGFace and CASIA-WebFace—both well-established in face verification literature. This serves as an ablation study to evaluate the impact of different visual backbones.
For the language modality, we use a GPT-2-based language model, pretrained on a large corpus of text. This provides a strong linguistic foundation before integrating visual features.

\textbf{2. Multimodal Mapper} Training Once unimodal training is complete, we freeze both the vision encoder and the language model, and train the cross-projection and image-projection layers. The goal of this stage is to align the visual representation space with the language model’s embedding space, allowing the model to condition textual generation on visual input effectively.

\textbf{3. End-to-End Fine-tuning:} In the final stage, we unfreeze all components of the model and perform end-to-end fine-tuning. This allows the entire architecture to jointly adapt, leading to a more nuanced integration of visual and linguistic features. We adopt a low learning rate with warm-up scheduling to ensure stable convergence and prevent catastrophic forgetting during this sensitive phase of training.

\subsection*{5.2 Loss Function}

To encourage more descriptive and diverse outputs from our model, we employ a variant of the standard cross-entropy loss, augmented with an entropy-based regularization term. This combined objective—referred to as \textit{Diversity Loss}—strikes a balance between accurate generation and lexical variability, which is especially important in multimodal generative tasks involving free-form language.

Let $\mathbf{z} \in \mathbb{R}^{B \times T \times V}$ be the predicted logits from the language model, where $B$ is the batch size, $T$ is the sequence length, and $V$ is the vocabulary size. The corresponding ground-truth target sequence is denoted by $\mathbf{y} \in \mathbb{N}^{B \times T}$.

The total loss $\mathcal{L}$ is defined as:
\[
\mathcal{L} = \mathcal{L}_{\text{CE}} - \lambda \cdot \mathcal{H}
\]
where:
\begin{itemize}
    \item $\mathcal{L}_{\text{CE}}$ is the standard cross-entropy loss between the predicted token distributions and the ground-truth targets.
    \item $\mathcal{H}$ is the mean entropy of the predicted token distributions.
    \item $\lambda$ is a scalar coefficient that controls the weight of the entropy regularization term.
\end{itemize}

The entropy $\mathcal{H}$ is computed as:
\[
\mathcal{H} = -\frac{1}{B \cdot T} \sum_{b=1}^{B} \sum_{t=1}^{T} \sum_{v=1}^{V} p_{b,t,v} \log(p_{b,t,v} + \epsilon)
\]
where $p_{b,t,v}$ represents the softmax-normalized probability of token $v$ at time step $t$ for batch $b$, and $\epsilon$ is a small constant added for numerical stability.

By subtracting the entropy term from the loss, we explicitly encourage the model to assign higher entropy to its output distributions—i.e., to be less confident and more exploratory. This regularization helps mitigate overconfident, generic outputs (e.g., repetitive or templated responses), pushing the model toward producing more expressive and contextually grounded text.

Empirically, we find that this diversity-promoting mechanism improves the fluency and specificity of generated descriptions, particularly in the later stages of end-to-end fine-tuning where overfitting to deterministic language patterns is more likely.

%% file: documents/results.tex
\subsection{Experimental Setup}

We compare our proposed architecture, \textbf{VerLM}, with two baselines: the \textbf{OneDiff} architecture and a \textbf{Base model} comprising a verification encoder, a mapper, and a language model. To ensure a fair comparison focused solely on the impact of the mapper architecture, we fix the backbone components across all models—using \texttt{GPT-2 Base} as the language model and \texttt{VGGFace} as the verification encoder.

In the \textbf{OneDiff} architecture, a cross-modal projector is introduced before the language model to align the image pairs with their corresponding textual descriptions. The \textbf{Base model} lacks this projector and serves as a minimal multimodal pipeline. Our proposed \textbf{VerLM} architecture extends this setup by incorporating a \textit{separator module} and a \textit{cross-projection layer} that enhances the multimodal alignment prior to language generation.

All three architectures were trained using the same training strategy to ensure a fair ablation study. The mapper components were initially trained for 30 epochs using the \texttt{Adam} optimizer with a learning rate of $1 \times 10^{-4}$. We employed the \texttt{Cosine Annealing with Warm Restarts} learning rate scheduler and a batch size of 64. After this stage, the entire model was fine-tuned end-to-end by unfreezing all modules and continuing training with a reduced learning rate of $1 \times 10^{-5}$, while keeping all other training configurations the same.

All experiments were conducted using \texttt{NVIDIA V100} GPUs.

\subsection{Evaluation Metrics and Performance Analysis}

\begin{table}[h]
\centering
\resizebox{\textwidth}{!}{
\begin{tabular}{lccc|ccc}
\toprule
\multirow{2}{*}{\textbf{Ablation Type}} 
& \multicolumn{3}{c|}{\textbf{Dataset 1}} 
& \multicolumn{3}{c}{\textbf{Dataset 2}} \\
\cmidrule(lr){2-4} \cmidrule(lr){5-7}
& \textbf{METEOR} & \textbf{BLEU} & \textbf{BERT} 
& \textbf{METEOR} & \textbf{BLEU} & \textbf{BERT} \\
\midrule
Our Model   & \textbf{0.3986} & \textbf{0.1557} & \textbf{0.9039} 
            & \textbf{0.3548} & \textbf{0.1338} & \textbf{0.9004} \\
OneDiff     & 0.3227 & 0.1143 & 0.8728 
            & 0.3419 & 0.1301 & 0.8980 \\
Base Model    & 0.3343 & 0.1052 & 0.8882 
            & 0.3306 & 0.1137 & 0.8955 \\
\bottomrule
\end{tabular}
}
\caption{Comparison of models across Dataset 1 and Dataset 2 using METEOR, BLEU, and BERTScore.}
\label{tab:ablations}
\end{table}

To assess the quality of generated descriptions across different models, we evaluate performance using three widely-adopted metrics: METEOR, BLEU, and BERTScore. Each of these metrics captures different aspects of generation quality, offering a comprehensive evaluation of fluency, precision, and semantic similarity.

\textbf{METEOR} (Metric for Evaluation of Translation with Explicit ORdering) computes a weighted F-score based on the alignment between the generated and reference sentences. It incorporates stemming, synonymy matching (using WordNet), and exact matches, making it more semantically aware than traditional n-gram overlap metrics. Higher METEOR scores indicate better alignment and semantic similarity with reference sentences.

\textbf{BLEU} (Bilingual Evaluation Understudy) measures the n-gram precision between the generated and reference text, typically up to 4-grams. While it is a widely used metric in machine translation and text generation, BLEU is known to be overly sensitive to exact word matches and may not capture semantic equivalence well, especially in open-ended generation tasks.

\textbf{BERTScore} leverages contextual embeddings from pretrained BERT models to compute a token-level similarity between candidate and reference sentences. Unlike BLEU and METEOR, BERTScore is semantic in nature and is capable of capturing paraphrastic similarity and richer contextual meaning.

\textbf{Performance Analysis.} As shown in Table~\ref{tab:ablations}, our proposed model \textbf{VerLM} consistently outperforms both OneDiff and Base Model across all three evaluation metrics on both datasets. Specifically, VerLM achieves the highest METEOR (0.3986 and 0.3548), BLEU (0.1557 and 0.1338), and BERTScore (0.9039 and 0.9004) on Dataset 1 and Dataset 2, respectively. These results suggest that VerLM generates descriptions that are not only more fluent and accurate at the surface level (as indicated by BLEU and METEOR), but also semantically richer and more contextually aligned with the reference (as evidenced by BERTScore).

In contrast, OneDiff and base Model demonstrate lower scores across all metrics, indicating that their generated outputs are less aligned with the ground truth both lexically and semantically. The consistent improvement of VerLM across datasets and metrics highlights the effectiveness of its cross-projection and separator modules in enhancing multimodal alignment and guiding the language model to produce higher-quality descriptions.

%% file: documents/ablations.tex
We implement our VerLM model for the task of explainable face verification using natural language. We begin by detailing the datasets used for training and evaluating the model in this setting. Following this, we describe the architectural design and training configuration of the proposed system. To validate its effectiveness, we benchmark our model against existing state-of-the-art approaches. Finally, we conduct an ablation study to analyze the impact of key architectural components and training strategies, highlighting the design choices that contribute most to model performance.

\subsection{Training Strategy}
 
\begin{table}[h]
\centering
\resizebox{\textwidth}{!}{
\begin{tabular}{lccc|ccc}
\toprule
\multirow{2}{*}{\textbf{Ablation Type}} 
& \multicolumn{3}{c|}{\textbf{Dataset 1}} 
& \multicolumn{3}{c}{\textbf{Dataset 2}} \\
\cmidrule(lr){2-4} \cmidrule(lr){5-7}
& \textbf{METEOR} & \textbf{BLEU} & \textbf{BERT} 
& \textbf{METEOR} & \textbf{BLEU} & \textbf{BERT} \\
\midrule
Train Mapper                  & 0.3672 & 0.1421 & 0.9015 
                             & 0.3519 & 0.1321 & 0.9001 \\
Train Entire End-to-End      & 0.3753 & 0.1516 & 0.9021 
                             & 0.3478 & 0.1309 & 0.8901 \\
Train Mapper Then Finetune   & \textbf{0.3986} & \textbf{0.1557} & \textbf{0.9039} 
                             & \textbf{0.3548} & \textbf{0.1338} & \textbf{0.9004} \\
\bottomrule
\end{tabular}
}
\caption{Ablation study comparing different training strategies on Dataset 1 and Dataset 2 using METEOR, BLEU, and BERTScore.}
\label{tab:training_ablation}
\end{table}

We explored three distinct training strategies for our proposed \textbf{VerLM} architecture to determine the most effective approach for aligning visual features with the language model. Each strategy was designed to assess the impact of different levels of parameter freezing and optimization flow on performance.

The first strategy involved training only the \textit{vision projection layer} and \textit{cross projection layer}, while keeping both the verification encoder and the language model frozen. These two projection modules are collectively referred to as the \textbf{mapper}, as they serve to bridge the vision and language domains. This setup aims to ensure that the visual features are mapped into the language model's embedding space without disrupting pretrained representations in the encoder or decoder.

The second strategy involved fully \textit{end-to-end training} of the entire model from the beginning. In this case, no components were frozen, allowing the optimization process to jointly adapt the verification encoder, mapper, and language model to the task. While more flexible, this method introduces challenges in stability and may require careful learning rate scheduling to prevent overfitting or forgetting.

The final and most effective approach was a \textbf{three-stage training procedure}. In this setup, we first train the mapper alone (as in the first strategy), then unfreeze the entire model and perform end-to-end fine-tuning. This progressive learning approach allows the mapper to first align the modalities without interference, after which the entire model is jointly optimized for improved multimodal integration.

As shown in Table \ref{tab:training_ablation}, the third strategy consistently outperformed the other two across both datasets and all evaluation metrics. Notably, it achieved the highest scores in METEOR (0.3986 and 0.3548), BLEU (0.1557 and 0.1338), and BERTScore (0.9039 and 0.9004). These results validate the effectiveness of a staged training paradigm in preserving pretrained representations while still allowing for holistic adaptation during fine-tuning.

\subsection{Scaling of Language Model}

\begin{table}[h]
\centering
\resizebox{0.75\textwidth}{!}{
\begin{tabular}{lccc}
\toprule
\textbf{Ablation Type} & \textbf{METEOR} & \textbf{BLEU} & \textbf{BERT} \\
\midrule
GPT Base   & 0.3672 & 0.1421 & 0.9015 \\
GPT Medium & 0.3937 & 0.1570 & 0.9036 \\
GPT Large  & \textbf{0.4098} & \textbf{0.1572} & \textbf{0.9047} \\
\bottomrule
\end{tabular}
}
\caption{Ablation study on the effect of language model size (GPT variants) on description quality, evaluated using METEOR, BLEU, and BERTScore.}
\label{tab:gpt_ablation}
\end{table}

Recent studies have shown that the performance of language models—and by extension, vision-language models—tends to improve with increasing model scale, often following a predictable power-law behavior \cite{kaplan2020scaling, hoffmann2022training}. To investigate this trend within our framework, we conducted an ablation study using three variants of the GPT-2 architecture: GPT-2 Base (124M parameters), GPT-2 Medium (355M), and GPT-2 Large (774M) \cite{radford2019language}.

As presented in Table~\ref{tab:gpt_ablation}, we observed a consistent improvement in performance across all evaluation metrics—METEOR, BLEU, and BERTScore—as the size of the language model increased. Specifically, GPT-2 Large achieved the highest scores, with a METEOR of 0.4098, BLEU of 0.1572, and BERTScore of 0.9047. These results align with prior findings and reinforce the idea that larger language models possess greater capacity to generate contextually rich and semantically aligned textual descriptions when paired with visual inputs.

\subsection{Model Strategy}

\begin{table}[h]
\centering
\resizebox{\textwidth}{!}{
\begin{tabular}{lccc|ccc}
\toprule
\multirow{2}{*}{\textbf{Ablation Type}} 
& \multicolumn{3}{c|}{\textbf{Dataset 1}} 
& \multicolumn{3}{c}{\textbf{Dataset 2}} \\
\cmidrule(lr){2-4} \cmidrule(lr){5-7}
& \textbf{METEOR} & \textbf{BLEU} & \textbf{BERT} 
& \textbf{METEOR} & \textbf{BLEU} & \textbf{BERT} \\
\midrule
With SEP                     & \textbf{0.3672} & \textbf{0.1421} & \textbf{0.9015} 
                             & \textbf{0.3519} & 0.1321          & 0.9001 \\
Without SEP                  & 0.3532          & 0.1308          & 0.8921 
                             & 0.3509          & \textbf{0.1323} & \textbf{0.9025} \\
Without Cross Projection     & 0.3343          & 0.1052          & 0.8882 
                             & 0.3398          & 0.1281          & 0.8871 \\
\bottomrule
\end{tabular}
}
\caption{Ablation study on model components: effect of separator (SEP) and cross-projection layer on performance across Dataset 1 and Dataset 2.}
\label{tab:sep_ablation}
\end{table}

We examine the impact of the two architectural components adapted from ADIFF - the insertion of a separator token (SEP) between the image embeddings and the use of a cross projection module. We compare three variants in this experiment: (1) \textit{With SEP} (the full VerLM model, including the SEP token and cross-projection layer), (2) \textit{Without SEP} (cross-projection module is retained but no explicit separator token is inserted), and (3) \textit{Without Cross-Projection} (both the SEP token and the entire cross-projection mechanism are removed, akin to a direct embedding prompt baseline). This experiment helps us evaluate how well the model can highlight the fine-grained facial differences when these design elements are ablated.

We found that incorporating the cross projection module with the separator token yields the best performance. The With Sep (i.e. full model) achieves the highest score in METEOR (0.3762), BLEU (0.142), AND BERTScore (0.9015) in dataset 1. Removing the separator token causes a slight drop in these metrics (METEOR falls by about 1.4 points, BLEU by $\sim$1 point), indicating that the SEP token provides a modest but consistent benefit. Notably, eliminating the entire cross-projection module leads to a significant performance degradation: METEOR plummets to 0.334, and BLEU drops to 0.105 for impostor pairs (a relative decrease of about 25\% and 26\% respectively, compared to the full model). BERTScore likewise declines (from $\sim$0.901 to 0.888), reflecting poorer semantic alignment with the reference explanations. These trends hold similarly for genuine (same-identity) pairs, underscoring the importance of the cross-projection design across both scenarios.

The superior performance with our full model suggests that VerLM's difference-focused strategy is effective. The cross-projection module evidently helps the model isolate and represent distinguishing attributes of the two faces in the latent space, thereby facilitating more precise difference descriptions. This aligns with observations from ADIFF, where a cross-projection layer (with a learned separator) was critical to “store difference attributes” in the prompt for the language model \cite{deshmukh2025adiff}. By inserting a dedicated token between the two face embeddings, VerLM explicitly demarcates the two inputs, enabling the transformer to attend to each face’s features without conflation. The slight performance dip without SEP suggests that even a single learned boundary token can improve the model’s ability to discern which visual features belong to which face, likely reducing ambiguity during comparison.In contrast, removing the entire cross-projection mechanism forces the language model to rely on a naive concatenation of image features (similar to the baseline in ADIFF or a simple prefix-tuning approach), which dramatically hampers its ability to articulate fine-grained differences. This result is consistent with prior vision-language models that emphasize structured multimodal fusion: for example, OneDiff \cite{hu2024onediffgeneralistmodelimage} introduces a “Visual Delta Module” to explicitly capture image differences, and ablations in that work show performance drops when the module is removed. Our findings here similarly demonstrate that a targeted difference-computation layer (and its SEP token) substantially boosts descriptive accuracy. 

\subsection{Textual Encoder Strategy}
\begin{table}[h]
\centering
\resizebox{\textwidth}{!}{
\begin{tabular}{lccc|ccc}
\toprule
\multirow{2}{*}{\textbf{Ablation Type}} 
& \multicolumn{3}{c|}{\textbf{Dataset 1}} 
& \multicolumn{3}{c}{\textbf{Dataset 2}} \\
\cmidrule(lr){2-4} \cmidrule(lr){5-7}
& \textbf{METEOR} & \textbf{BLEU} & \textbf{BERT} 
& \textbf{METEOR} & \textbf{BLEU} & \textbf{BERT} \\
\midrule
Without Text Encoder & 0.3544 & 0.1363 & 0.8992 
                     & 0.3393 & 0.1218 & 0.8901 \\
With Text Encoder    & \textbf{0.3672} & \textbf{0.1421} & \textbf{0.9015} 
                     & \textbf{0.3519} & \textbf{0.1321} & \textbf{0.9001} \\
\bottomrule
\end{tabular}
}
\caption{Ablation study on the impact of including a text encoder in the VerLM architecture. Results reported on Dataset 1 and Dataset 2 using METEOR, BLEU, and BERTScore.}
\label{tab:text_encoder_ablation}
\end{table}

To assess the impact of incorporating a textual encoder, we conducted an ablation study comparing model performance with and without this component. In the full \textbf{VerLM} architecture, the text encoder processes the tokenized input prompts into semantic embeddings, which are then used in conjunction with visual embeddings during generation.

As shown in Table~\ref{tab:text_encoder_ablation}, the inclusion of the text encoder results in consistent improvements across all metrics—METEOR, BLEU, and BERTScore—on both datasets. These results highlight the critical role of the text encoder in preserving semantic alignment and coherence between the input prompt and the generated output.

We hypothesize that removing the text encoder introduces a misalignment between the textual conditioning and the generated language, thereby degrading output quality. This observation is consistent with recent findings in multimodal generative modeling \cite{deshmukh2024pengiaudiolanguagemodel}, which emphasize the importance of a dedicated text encoder for maintaining contextual consistency and improving generation fidelity.

\subsection{Different Image Encoders}
\begin{table}[h]
\centering
\resizebox{\textwidth}{!}{
\begin{tabular}{lccc|ccc}
\toprule
\multirow{2}{*}{\textbf{Image Encoder}} 
& \multicolumn{3}{c|}{\textbf{Dataset 1}} 
& \multicolumn{3}{c}{\textbf{Dataset 2}} \\
\cmidrule(lr){2-4} \cmidrule(lr){5-7}
& \textbf{METEOR} & \textbf{BLEU} & \textbf{BERT} 
& \textbf{METEOR} & \textbf{BLEU} & \textbf{BERT} \\
\midrule
VGGFace & \textbf{0.3672} & \textbf{0.1421} & \textbf{0.9015} 
        & \textbf{0.3519} & \textbf{0.1321} & \textbf{0.9001} \\
CASIA   & 0.3413          & 0.1389          & 0.8908  
        & 0.3209          & 0.1203          & 0.887 \\
\bottomrule
\end{tabular}
}
\caption{Comparison of image encoders (VGGFace vs. CASIA-WebFace) on Dataset 1 and Dataset 2 using METEOR, BLEU, and BERTScore.}
\label{tab:image_encoder_ablation}
\end{table}
To assess the importance of the visual encoder, we conduct an ablation study by swapping VerLM’s face feature extractor. Specifically, we compare two widely-used face recognition models: \textit{VGGFace}, trained on the VGGFace dataset \cite{Parkhi15b}, and \textit{CASIA}, a ResNet-based model trained on the CASIA-WebFace dataset \cite{yi2014learningfacerepresentationscratch}. Both encoders produce latent embeddings from facial images, which are then passed to the model's cross-projection module to condition the language generation process.

As shown in Table~\ref{tab:image_encoder_ablation}, the model using VGGFace significantly outperforms the CASIA-based model across METEOR, BLEU, and BERTScore metrics on both datasets. The VGGFace encoder achieves higher semantic and syntactic alignment with reference descriptions, suggesting that it provides more discriminative and expressive facial embeddings.

We hypothesize that this performance gap arises not only from architectural differences but also from the underlying training datasets. Since our model is trained using facial inputs aligned with the VGGFace embedding space, using an encoder trained on the same dataset (VGGFace) results in better modality alignment. In contrast, CASIA-WebFace differs in its identity distribution, pose diversity, and image quality, which likely introduces a domain mismatch and degrades the downstream generation performance.

This finding underscores the importance of dataset alignment between the pretrained vision backbone and the multimodal training regime, especially in tasks involving fine-grained identity-level conditioning.

\subsection{Model Scaling}
\begin{table}[h]
\centering
\resizebox{\textwidth}{!}{
\begin{tabular}{lccc|ccc}
\toprule
\multirow{2}{*}{\textbf{Number of GPT Layers}} 
& \multicolumn{3}{c|}{\textbf{Dataset 1}} 
& \multicolumn{3}{c}{\textbf{Dataset 2}} \\
\cmidrule(lr){2-4} \cmidrule(lr){5-7}
& \textbf{METEOR} & \textbf{BLEU} & \textbf{BERT} 
& \textbf{METEOR} & \textbf{BLEU} & \textbf{BERT} \\
\midrule
2  & 0.3403 & 0.1241 & 0.8975 
    & 0.3329 & 0.1221 & 0.8891 \\
4  & 0.3672 & 0.1421 & 0.9015 
    & 0.3519 & 0.1321 & 0.9001 \\
8  & 0.3662 & 0.1317 & 0.8990 
    & 0.3532 & 0.1317 & 0.9015 \\
16 & \textbf{0.3703} & \textbf{0.1486} & \textbf{0.9094} 
    & \textbf{0.3669} & \textbf{0.1415} & \textbf{0.9033} \\
\bottomrule
\end{tabular}
}
\caption{Ablation study on the number of transformer layers in GPT-base. Results are reported on Dataset 1 and Dataset 2 using METEOR, BLEU, and BERTScore.}
\label{tab:num_gpt_layers_ablation}
\end{table}

%% file: documents/conclusion.tex
In this work, we introduced VerLM, a difference-aware vision–language architecture for explainable face verification that generates natural-language rationales grounded in paired facial embeddings. Under a controlled comparison where the language model (GPT-2 Base) and face encoder (VGGFace) are held fixed, VerLM consistently outperforms both a minimal multimodal baseline and the OneDiff-inspired alternative across METEOR, BLEU, and BERTScore on two datasets, demonstrating more fluent text and stronger semantic alignment with references. Our ablations highlight that these gains come from explicit difference modeling, not simply adding capacity. A staged training procedure—training the mapper first, then fine-tuning end-to-end yields the best overall results, indicating that progressive alignment helps preserve pretrained representations while still enabling holistic adaptation.  

Architecturally, the cross-projection module is the primary driver of explanation quality, while the separator token (SEP) provides a smaller but consistent improvement by preventing feature conflation between the two faces.  We also find that explanation quality improves with larger language models and deeper decoders, and that both a dedicated text encoder and better aligned face embeddings (VGGFace vs. CASIA) materially impact performance underscoring the importance of multimodal alignment at every stage of the pipeline. Looking ahead, promising directions include scaling to stronger instruction-tuned LLMs, expanding to more diverse face verification conditions (pose/occlusion/lighting), adding human centered evaluation for explanation usefulness and faithfulness, and tightening visual grounding (e.g., coupling generated language with region-level evidence) so that explanations are not only fluent, but also reliably attributable to the true factors driving verification decisions.

%% file: documents/appendix.tex
\footnotesize
\begin{center}
\begin{longtable}{ |c|c|c|c| } 
 \hline
 Image 1 & Image 2 & Llama 3.2 Description & Our Model Results \\ 
 \hline
 \includegraphics[scale=0.25]{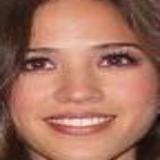} & \includegraphics[scale=0.25]{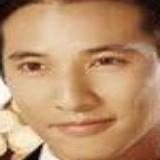} &\begin{varwidth}{15em}the two facial images share several matching features, including a heart-shaped face, almond-shaped dark brown eyes, straight nose, thin eyebrows, high cheekbones, small chin, and smooth forehead. however, notable differences include the jawline shape, with one image featuring a soft jawline and the other a slightly more defined jawline, and the eyebrow shape, with one image having arched eyebrows and the other straight eyebrows.\end{varwidth} & \begin{varwidth}{15em}the two facial images exhibit distinct differences in key identifying features. person 1 has a round face, almond-shaped eyes, and a straight nose, while person 2 has a heart-shaped face, almond-shaped eyes, and a straight nose. the notable discrepancies in facial structure, skin tone, and hair color suggest they are not the same individual \end{varwidth}\\ 
 \hline
 \includegraphics[scale=0.25]{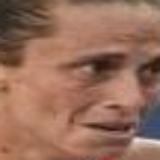} & \includegraphics[scale=0.25]{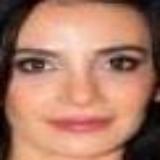} &\begin{varwidth}{15em}the two images show a woman with dark hair and fair, warm skin tone. the first image features a bun, while the second has shoulder-length hair with loose waves and dark brown eyes. the facial structure appears similar, but the hair style and eye color discrepancies suggest they are not the same individual. \end{varwidth}& \begin{varwidth}{15em}the two facial images exhibit distinct differences in key identifying features. person 1 has a fair skin tone, dark brown hair, and a defined jawline, while person 2 has a warm, golden brown skin tone, dark brown hair, and a defined jawline. the notable discrepancies in skin tone, hair color, and facial structure suggest they are not the same individual \end{varwidth}\\ 
 \hline
 \includegraphics[scale=0.25]{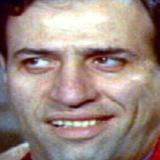} & \includegraphics[scale=0.25]{images/appendix_images/0058_01.jpg} &\begin{varwidth}{15em} the two facial images exhibit a high degree of similarity in key identifying features, including a rounded face shape, almond-shaped dark brown eyes, straight nose, thin mouth, and dark eyebrows. the main difference lies in the hair, with one image showing a more pronounced wave than the other.\end{varwidth}& \begin{varwidth}{15em} the two facial images exhibit numerous matching features, including oval face shape, almond-shaped dark brown eyes, straight nose, thick eyebrows, and fair skin tone. the only minor difference is the slight variation in the hair color and style, which is not significant enough to suggest they are not the same individual. \end{varwidth}\\ 
 \hline
 \includegraphics[scale=0.25]{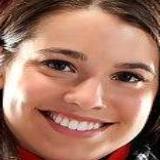} & \includegraphics[scale=0.25]{images/appendix_images/0065_01.jpg} &\begin{varwidth}{15em} the two facial images exhibit a high degree of similarity. matching features include an oval face shape, almond-shaped dark brown eyes, and fair skin tone. the overall facial structure is identical, confirming that the images represent the same individual.\end{varwidth}& \begin{varwidth}{15em} the two facial images exhibit numerous matching features, including oval face shape, almond-shaped dark brown eyes, straight nose, thick eyebrows, and fair skin tone. the only minor difference is the presence of a small mole in the first image, but this is insignificant in comparison to the overall similarity.\end{varwidth}\\ 
 \hline
 \includegraphics[scale=0.25]{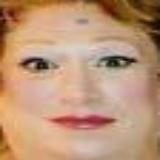} & \includegraphics[scale=0.25]{images/appendix_images/0270_02.jpg} &\begin{varwidth}{15em} the two facial images share several key features, including fair skin, oval-shaped face, almond-shaped dark brown eyes, and thin, arched eyebrows. the notable similarities suggest that these images represent the same individual.\end{varwidth}& \begin{varwidth}{15em} the two facial images exhibit numerous matching features, including oval face shape, almond-shaped dark brown eyes, straight nose, thin eyebrows, and fair skin tone. the only minor difference is the slight variation in hair color, but this is negligible in comparison to the overall similarity.\end{varwidth}\\ 
 \hline
 \includegraphics[scale=0.25]{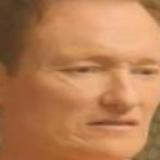} & \includegraphics[scale=0.25]{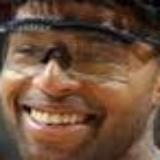} &\begin{varwidth}{15em} the two images exhibit distinct differences in key identifying features. the second image shows a man with a medium-brown skin tone, short beard and mustache, black dreadlocks, almond-shaped dark brown eyes, and oval face shape. the lack of discernible features in the first image and significant differences in the second image suggest they are not the same individual.\end{varwidth}& \begin{varwidth}{15em} the two facial images exhibit distinct differences in key identifying features. person 1 has a fair skin tone, dark brown eyes, and a prominent jawline, while person 2 has a medium to dark brown skin tone, almond-shaped eyes, and a straight nose. the notable discrepancies in skin tone, eye color, and facial structure suggest they are not the same individual.\end{varwidth}\\ 
 \hline
 \includegraphics[scale=0.25]{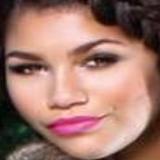} & \includegraphics[scale=0.25]{images/appendix_images/0277_01.jpg} &\begin{varwidth}{15em} the two facial images share several matching features, including skin tone, long dark hair, jawline, facial structure, and eye color. however, a difference in facial expression exists, with one image showing a serious expression.\end{varwidth}& \begin{varwidth}{15em} the two facial images exhibit numerous matching features, including oval face shape, almond-shaped dark brown eyes, straight nose, thick eyebrows, and fair skin tone. the only minor difference is the presence of a small mole in the first image, but this is insignificant in comparison to the overall similarity.\end{varwidth}\\ 
 \hline
 \includegraphics[scale=0.25]{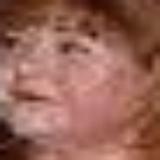} & \includegraphics[scale=0.25]{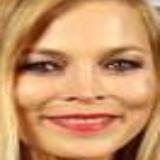} &\begin{varwidth}{15em} the two facial images share several key features, including oval face shape, almond-shaped dark brown eyes, straight eyebrows, high cheekbones, and smooth skin tone. however, notable differences include hair color, style, eyebrow color, and jawline shape. despite these differences, the facial structure and eye shape remain consistent. based on these findings, the two images do not represent the same individual.\end{varwidth}& \begin{varwidth}{15em} the two facial images exhibit distinct differences in key identifying features. person 1 has a fair skin tone, blonde hair, and a defined jawline, while person 2 has a fair skin tone, dark brown hair, and a defined jawline. the notable discrepancies in skin tone, hair color, and facial structure suggest they are not the same individual.\end{varwidth}\\ 
 \hline
 \includegraphics[scale=0.25]{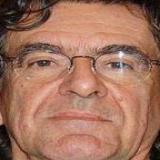} & \includegraphics[scale=0.25]{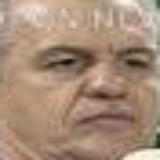} &\begin{varwidth}{15em} the two images exhibit distinct differences in facial structure and hair color. the first image depicts a man with a square-shaped face, almond-shaped dark brown eyes, and thick dark eyebrows. in contrast, the second image shows a man with short gray hair and no matching facial features. the drastic change in hair color suggests they are not the same individual.\end{varwidth}& \begin{varwidth}{15em} the two facial images exhibit distinct differences. person 1 has a round face shape, prominent jawline, and dark brown eyes, while person 2 has a round face shape, prominent jawline, and dark brown eyes. the notable discrepancies in facial structure, jawline, and eye color suggest they are not the same individual.\end{varwidth}\\ 
 \hline
 \includegraphics[scale=0.25]{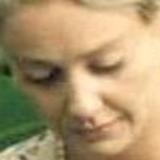} & \includegraphics[scale=0.25]{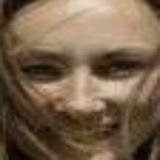} &\begin{varwidth}{15em} the two facial images share several matching characteristics, including oval face shapes, dark brown almond-shaped eyes, straight noses, and prominent cheekbones. however, the first image shows a woman with short blonde hair, a smoother complexion, and a less angular jawline, while the second image depicts a person with long straight light brown hair, fine lines, and a stronger jawline. based on these distinct differences, the two images do not represent the same individual.\end{varwidth}& \begin{varwidth}{15em} the two facial images exhibit distinct differences in key identifying features. person 1 has a fair skin tone, almond-shaped eyes, and a straight nose, while person 2 has a light skin tone, almond-shaped eyes, and a straight nose. the notable discrepancies in skin tone, eye color, and nose shape suggest they are not the same individual.\end{varwidth}\\ 
 \hline
  \includegraphics[scale=0.25]{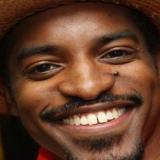} & \includegraphics[scale=0.25]{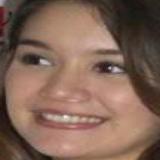} &\begin{varwidth}{15em} the two images show distinct individuals. person 1 is a black man with a rounded face, almond-shaped eyes, and a prominent nose. person 2 is a woman with long brown hair, a white top, and a different facial structure. the key differences indicate they are not the same person.\end{varwidth}& \begin{varwidth}{15em} the two facial images exhibit distinct differences in key identifying features. person 1 has a fair skin tone, almond-shaped dark brown eyes, and a straight nose, while person 2 has a medium to dark brown skin tone, almond-shaped dark brown eyes, and a straight nose. the notable discrepancies in skin tone, eye color, and nose shape suggest they are not the same individual.\end{varwidth}\\ 
 \hline
 \includegraphics[scale=0.25]{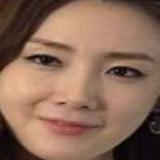} & \includegraphics[scale=0.25]{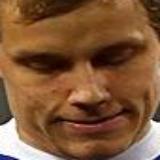} &\begin{varwidth}{15em} the facial images of person 1 and person 2 exhibit key differences. person 1 has a brown skin tone, long brown hair, while person 2 has a fair skin tone, short light-brown hair, and an oval face shape with almond-shaped eyes. the discrepancies in skin tone, hair color and style, and facial structure suggest they are different individuals.\end{varwidth}& \begin{varwidth}{15em} the two facial images exhibit distinct differences in key identifying features. person 1 has a fair skin tone, almond-shaped eyes, and a straight nose, while person 2 has a light skin tone, almond-shaped eyes, and a straight nose. the notable discrepancies in skin tone, eye color, and nose shape suggest they are not the same individual.\end{varwidth}\\ 
 \hline
 \includegraphics[scale=0.25]{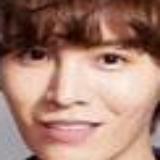} & \includegraphics[scale=0.25]{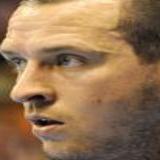} &\begin{varwidth}{15em} the two images exhibit distinct differences in key identifying features. person 1 has a round face, fair skin, and dark brown hair, while person 2 has an oval face, light to medium brown skin, and light blond hair. the facial features, such as nose shape, eyebrow shape, and jawline, also differ significantly.\end{varwidth}& \begin{varwidth}{15em} the two facial images exhibit distinct differences. person 1 has a round face, almond-shaped eyes, and a straight nose, while person 2 has a heart-shaped face, almond-shaped eyes, and a straight nose. the notable discrepancies in facial structure, skin tone, and hair color suggest they are not the same individual.\end{varwidth}\\ 
 \hline
 \includegraphics[scale=0.25]{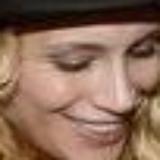} & \includegraphics[scale=0.25]{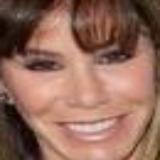} &\begin{varwidth}{15em} the two facial images share some matching features, including fair skin tone, small nose, prominent cheekbones, and defined jawline. however, they differ in hairstyles and colors, eye positions, and eyebrow shapes, suggesting the images are not of the same individual.\end{varwidth}& \begin{varwidth}{15em} the two facial images exhibit distinct differences in key identifying features. person 1 has a fair skin tone, dark brown hair, and a defined jawline, while person 2 has a warm, golden brown skin tone, dark brown hair, and a defined jawline. the notable discrepancies in skin tone, hair color, and facial structure suggest they are not the same individual.\end{varwidth}\\ 
 \hline
 \includegraphics[scale=0.25]{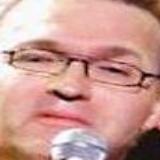} & \includegraphics[scale=0.25]{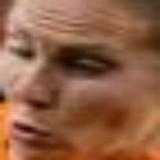} &\begin{varwidth}{15em} the two facial images exhibit distinct differences in key features. the first depicts a woman with a round face, soft jawline, brown or hazel eyes, and a non-prominent nose, while the second shows a woman with a flat face, prominent cheekbones, dark eyes, and a defined jawline. the notable discrepancies suggest that they are not the same individual.\end{varwidth}& \begin{varwidth}{15em} the two facial images exhibit distinct differences in key identifying features. person 1 has a fair skin tone, almond-shaped eyes, and a straight nose, while person 2 has a light skin tone, almond-shaped eyes, and a pointed nose. the notable discrepancies in skin tone, eye color, and nose shape suggest they are not the same individual.\end{varwidth}\\ 
 \hline
 \includegraphics[scale=0.25]{images/appendix_images/0104_01.jpg} & \includegraphics[scale=0.25]{images/appendix_images/0104_01.jpg} &\begin{varwidth}{15em} the two facial images depict a bald man with a prominent forehead, high cheekbones, strong jawline, light tan skin tone, and piercing blue eyes. the only minor difference lies in the background, but the facial features remain identical, suggesting they belong to the same individual.\end{varwidth}& \begin{varwidth}{15em} the two facial images exhibit numerous matching features, including a prominent forehead, almond-shaped dark brown eyes, straight nose, thick eyebrows, and fair skin tone. the only minor difference is the slight variation in the hair color and style, which is not significant enough to suggest they are not the same individual.\end{varwidth}\\ 
 \hline
 \includegraphics[scale=0.25]{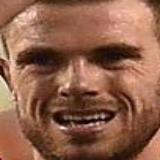} & \includegraphics[scale=0.25]{images/appendix_images/0539_01.jpg} &\begin{varwidth}{15em} the two facial images exhibit similarities in several key features, including short brown hair, straight noses, dark brown eyebrows, prominent cheekbones, defined jawlines, and fair skin tone. the matching features suggest that they belong to the same individual.\end{varwidth}& \begin{varwidth}{15em} the two facial images exhibit numerous matching features, including oval face shape, almond-shaped dark brown eyes, straight nose, thick eyebrows, and fair skin tone. the only minor difference is the slight variation in the hair color and style, which is not significant enough to suggest they are not the same individual.\end{varwidth}\\ 
 \hline
 \includegraphics[scale=0.25]{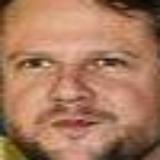} & \includegraphics[scale=0.25]{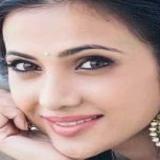} &\begin{varwidth}{15em} the two facial images exhibit similarities in face shape, forehead, cheekbones, nose, and eye shape. however, significant differences exist in skin tone, facial hair, hair style, mouth shape, and presence of fine lines.\end{varwidth}& \begin{varwidth}{15em} the two facial images exhibit distinct differences in key identifying features. person 1 has a fair skin tone, almond-shaped dark brown eyes, and a straight nose, while person 2 has a warm, golden brown skin tone, almond-shaped dark brown eyes, and a straight nose. the notable discrepancies in skin tone, eye color, and nose shape suggest they are not the same individual.\end{varwidth}\\ 
 \hline
  \includegraphics[scale=0.25]{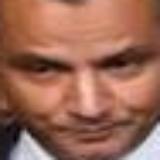} & \includegraphics[scale=0.25]{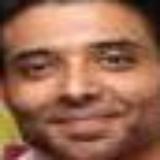} &\begin{varwidth}{15em} the facial features of both images exhibit several key differences. the most notable discrepancies include skin tone (medium to dark in person 1 vs. light in person 2), facial hair (stubble in person 1 vs. none in person 2), and jawline shape.\end{varwidth}& \begin{varwidth}{15em} the two facial images exhibit distinct differences. person 1 has a medium to dark brown skin tone, dark brown hair, and a prominent jawline, while person 2 has a light to medium brown skin tone, short dark hair, and a prominent jawline. the notable discrepancies in skin tone, hair color, and facial structure suggest they are not the same individual.\end{varwidth}\\ 
 \hline
  \includegraphics[scale=0.25]{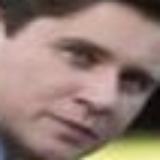} & \includegraphics[scale=0.25]{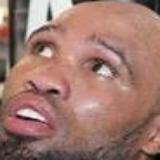} &\begin{varwidth}{15em} the two facial images exhibit distinct differences. the first is a blurry photograph of a man with short, dark hair, while the second is a close-up of a man with a shaved head and short beard. despite similar skin tone and facial structure, the significant differences in hair style, facial hair, and attire suggest they are not the same individual.\end{varwidth}& \begin{varwidth}{15em} the two facial images exhibit distinct differences in key identifying features. person 1 has a fair skin tone, almond-shaped eyes, and a straight nose, while person 2 has a dark skin tone, almond-shaped eyes, and a pointed chin. the notable discrepancies in skin tone, eye color, and facial structure suggest they are not the same individual.\end{varwidth}\\ 
 \hline
 \caption{20 image pairs with their descriptions and model results.} 
\end{longtable}
% needs to go inside longtable environment
\label{tab:appendixImages}
\end{center}